\documentclass[conference]{ieeetran}
\IEEEoverridecommandlockouts
\usepackage{cite}
\usepackage{amsmath,amssymb,amsfonts}
\usepackage{algorithmic}
\usepackage{graphicx}
\usepackage{textcomp}
\usepackage{xcolor}
\usepackage{booktabs}
\def\BibTeX{{\rm B\kern-.05em{\sc i\kern-.025em b}\kern-.08em
    T\kern-.1667em\lower.7ex\hbox{E}\kern-.125emX}}
\begin{document}


\title{A Distributed Consensus Particle Filter for Target Tracking using Autonomous Surface Vessels\\
\thanks{This work was funded by the NRL 6.1 Base Program}
\thanks{Distribution Statement A. Approved for public release: distribution is unlimited.}
}
\author{
    Carter Noh\textsuperscript{1}\thanks{\textsuperscript{1}Brigham Young University} \and
    Kyle Crandall\textsuperscript{2}\thanks{\textsuperscript{2}US Naval Research Lab, Code 5534} \and
    Connor Yates\textsuperscript{2} \and
    Corbin Wilhelmi\textsuperscript{2}
}

\maketitle

\begin{abstract} 
Maritime target tracking over large distances often requires multi-agent teams without centralized coordination, and intermittent communication.
Each agent must maintain an independent estimate that can take advantage of opportunistic communications availability when possible.
This can lead to overly confident local estimates in the absence of external data.
In this work, we propose an augmentation to a classical particle filter implementation that accounts for this potential source of error by forcing particles to spread strategically in the absence of informative updates from other sensor nodes.
We demonstrate our method using Unmanned Surface Vessels (USVs) on a lake, and show that our augmentations do not deteriorate nominal performance, and provide an advantage in some specific edge cases.
\end{abstract}

\section{Introduction}
Maritime tracking is key for a variety of tasks ranging form search and rescue, to law enforcement. 
This often requires multi-agent teams to coordinate without centralized control over vast ocean areas, where radio communication can be noisy and intermittent. 
Effective target tracking thus requires distributed solutions, where agents maintain independent estimates and take advantage of opportunistic communications when they are available.
Bayes theorem is the bedrock of such a distributed system, allowing the agent to update its estimates given new information from a variety of sources such as a model of the target's expected behavior, data from its own sensors, or information from other nodes in the system.
Particle filters are a common approach to approximate complex distributions using Bayesian updates.
Such filters have been used before for distributed tracking \cite{Hlinka2013, Ghirmai2016, Gao2026}, but often have stringent communications requirements.

Distributed consensus problems are a broader category of problems that have also been widely studied and utilized in communications disadvantaged situations \cite{Hadjicostis2024, Mehyar2007, Sorge2022, Macker2024}.
These kinds of algorithms share information opportunistically and update local estimates given new information.
However, they can also have stringent communications requirements, and do not perform well given large gaps in communication without additional heuristics.

We can draw some parallels from the distributed consensus problem where a prolonged gap in communications between different sets of nodes results in solving separate local problems, and combining the solutions when communications resume.
We extend this concept to the distributed Bayes filter problem where nodes can maintain local estimates when necessary, and coordinate updates with each-other when they can.
For such a system to work, it is important that the local estimates represent the true local distribution, however it easy for these filters to overly rely on prior updates, and yield a result that overly confident.

\begin{figure}[t]
	\centering
	\includegraphics[width=\linewidth]{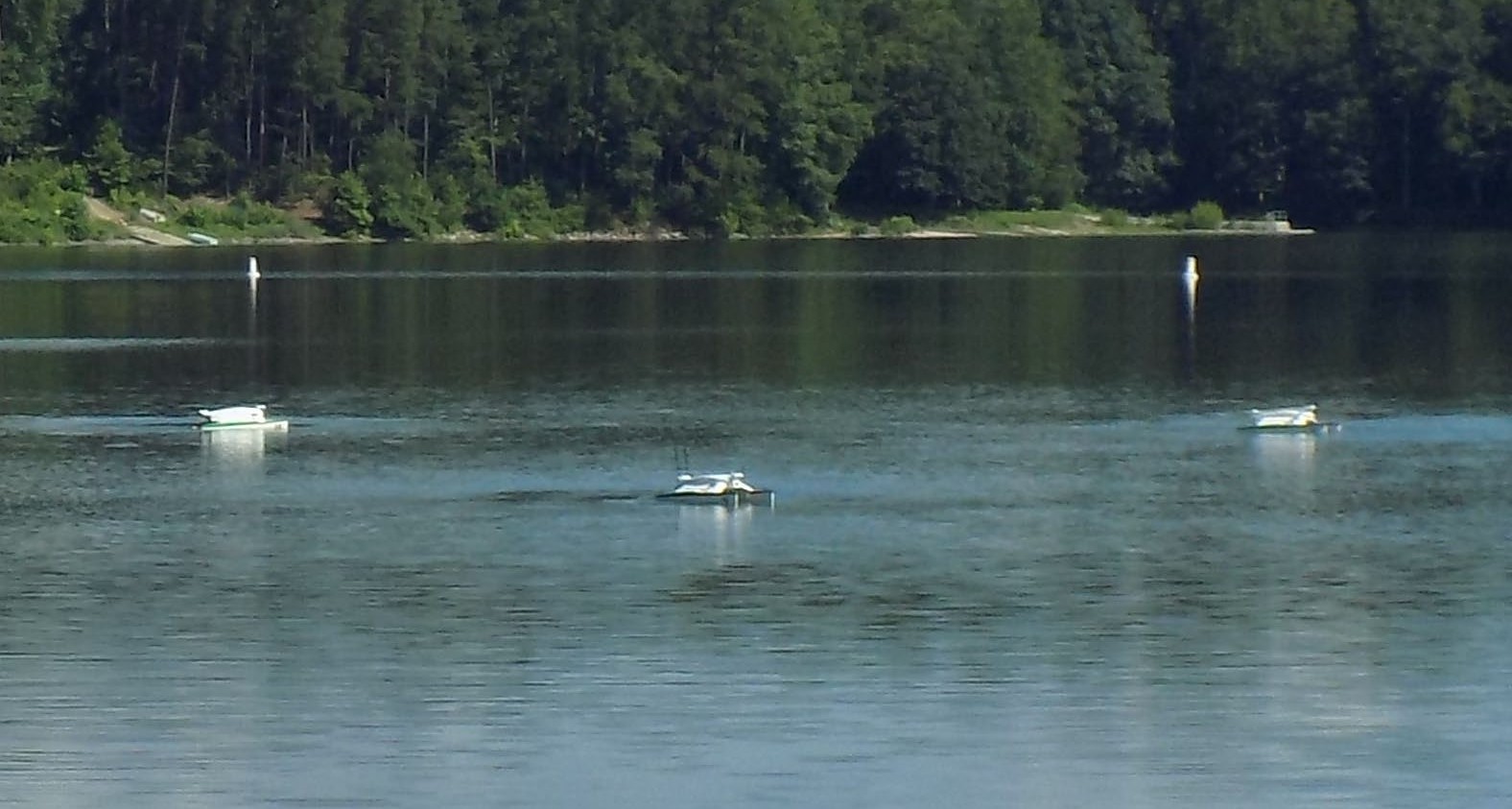}
	\caption{Two SeaRobotics HYCAT USVs use a distributed consensus particle filter with our proposed particle spreading heuristic to track the position of a third USV on a lake.}
	\label{fig:testing}
\end{figure}

In this work, we present a mechanism that actively attenuates the influence of prior updates and forces the local distribution to better approximate the recent local data in the absence of external updates.
This is done by spreading particles along the sensor manifold forcing them to distributed themselves along the the new probability distribution evenly until additional external data is received.
We demonstrate this technique in field tests on a team of Unmanned Surface Vehicles (USVs), as shown in Fig. \ref{fig:testing}. We show that in the nominal case, our proposed augmentations do not degrade the particle filter performance, and in certain edge cases, offer increased recovery speeds from drift due to communications dropout.

\section{Methods}

\subsection{Problem Formulation}
We consider two sets of USVs, a team of agents equipped with DoA sensors and one target emitting signal to be tracked by the agents.
Each agent estimates its own global position and orientation, and receive noisy DoA measurements of the target.
The agents within communication distance can exchange information, but communications are not guaranteed and subject to dropout.
The agents' objective is to estimate the position of the target over time.

The network of agents can be approximated as a communication graph, where each agent is a node that can communicate with its neighbors on the graph.
Connections between nodes change over time as agents move in and out of communication range or when communications are disrupted. 
If the graph connections remain stable for sufficient time, the model has guarantees of convergence through distributed average consensus (DAC) \cite{Hadjicostis2024, Mehyar2007, Sorge2022, Macker2024}.
We build on this concept to allow the agents to share their estimates with each other, but without having to share their measurements.
This results in a system where agents can take advantage of gossip style algorithms as each agent's estimate will incorporate the their own measurements, and any updates received from neighbors, eventually accounting for all measurements.
This is opposed to sharing measurements which would require forwarding values among the team and tracking which measurements have been applied and which have not which becomes cumbersome quickly.
This allows each agent to perform a Bayesian update to incorporate the new information.
We explore two mechanisms of doing this, a strict Bayesian approach where the world is tiled, and each tile holds a probability of the target being located in that tile as a baseline, and a distributed particle filter as our main focus.
In particular, we consider the consequences of periods of dropout in the communication between the agents and propose mechanisms to enhance recovery from the drift that such a disturbance can create in the estimates.

\subsection{Distributed Particle Filter}
The distributed particle filter is implemented by first having each agent track the target using a local particle filter. 
At time $t$, agent $i$ has a set of $M$ particles $\mathcal{P}^i_t = \{p^i_{t,1}, p^i_{t,2}, ... p^i_{t,M} \}$ , where each particle $p^i_{t,k} \in \mathbb{R}^2$ represents a possible location of the target.
this set of particles represents the ``likelihood'' distribution of the estimate.
Particles are updated in three ways: at every time step to account for predicted vehicle motion, when the agent receives a measurement, and when the agent receives an update from another agent.
These updates are performed using Bayes' rule.

\subsubsection{Prediction Update}
At each time step, the particles in $\mathcal{P}^i_t$ are propagated using a random walk to model of the target's predicted motion $T_i$:
\begin{equation} \label{eq:propagate}
p^i_{t+1} = p^i_t + T_idt
\end{equation}
Because the nature of the target may be unknown, $T$ is calculated by uniformly sampling a distance to move $r_i\in\left[0,r_{max}\right]$ and an angle $\beta_i\in\left[-\pi,\pi\right]$ that is the direction of movement:
\begin{equation} \label{eq:motionmodel}
T_i = \frac{r_i}{dt} \begin{bmatrix} \cos(\beta_i) \\ \sin(\beta_i) \\ \end{bmatrix}
\end{equation}
This sampling is done for each particle, so each particle is performing a unique update each step.

\subsubsection{Measurement Update}
For each particle $p^i_{t,k}$, the likelihood of a measurement $\theta$ is $w^i_{t,k} = \mathcal{N}(\theta; \hat{\theta}(p^i_{t,k}), \sigma^2)$.
When an agent receives a bearing measurement from the DoA sensor, it weights the particles using this likelihood function. 
The posterior estimate is then calculate by sampling from the set of particles using the normalized weights as probabilities forming a Bernoulli distribution.

\subsubsection{Consensus Update}
Like with the DAC solutions discussed previously, we can perform a similar update for particle filters allowing agents to share their estimates with each other, and performing local updates appropriately \cite{Hlinka2013, Ghirmai2016, Gao2026}. 
For our implementation, at each communication window agent $j$ shares a subset of its particles $\mathcal{P}^j_{t,s}$ with its neighbors.
Notably, this subset is a uniform sampling of the current particle set.
When agent $i$ receives particles $\mathcal{P}^j_{t,s}$, it gives each particle in $\mathcal{P}^j_{t,s}$ a weight $w_s = \frac{M}{M_j}$, where $M$ is the number of particles in $\mathcal{P}^i_t$ and $M_j$ is the number of particles in $\mathcal{P}^j_{t,s}$.
It then re-samples $\mathcal{P}^i_t$ from $\mathcal{P}^i_t \cup \mathcal{P}^j_{t,s}$ using the weights of the new particles, or 1 for current particles, to form a Bernoulli distribution similar to measurement update.

\subsection{Particle Spreading Heuristic}
Over time, the consensus updates between agents $i$ and $j$ drive the distributions $\Theta^i_t$ and $\Theta^j_t$ to converge and the particles $\mathcal{P}^i_t$ and $\mathcal{P}^j_t$ to cluster around a point. 
If inter-agent communication drops out for a period of time, data received from previous updates from other agents becomes increasingly irrelevant, and each agent can only constrain the distribution to its ``measurement manifold''. 
In the case of a DoA sensor, the measurement manifold is the line of bearing: the angle to the target is constrained, but the target's location along the line of bearing gains uncertainty as the dropout period progresses. 
To properly represent the uncertainty in the underlying distribution, the particles should spread out along the measurement manifold over time until a new consensus update is received, however the motion model alone may not be sufficient to cause the necessary spreading, leading the particles to remain tightly clustered and presenting a falsely certain estimate of the target's location.
If this happens and the agent receives a consensus update from another agent at the end of a dropout period, the resulting re-sampled distribution may be multi-modal, or otherwise deteriorated, resulting in a poor estimate that is difficult to recover from. 

To aid convergence after periods of inter-agent dropout, we introduce a heuristic that forces particles to spread along their measurement manifold quickly by applying an additional movement step during the prediction update. 
This extra movement is calculated by projecting the particles onto the measurement manifold, computing Voronoi cells from the particles, then moving each agent towards the center of its Voronoi cell.
This results in the particles distributing themselves evenly across the measurement manifold, resulting in decreased influence on the estimate from previous consensus updates.

\begin{figure}[t]
    \centering
    \includegraphics[width=\linewidth]{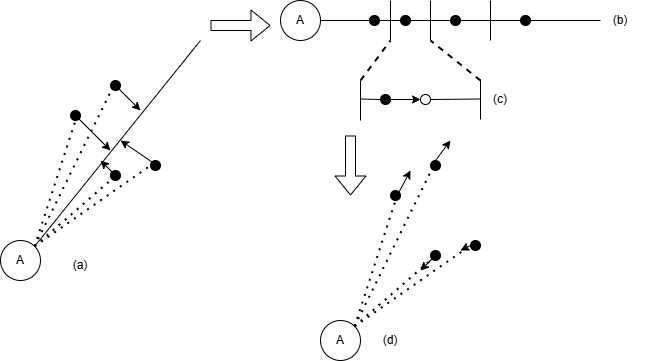}
    \caption{Particle spreading using Voronoi cells. (a) Particles are projected onto the mean sensor manifold. (b) The manifold is tessellated with Voronoi partitions, one for each particle. (c) The particles are moved towards the center of their respective Voronoi partitions. (d) This particle motion is projected back into the state space of the target.}
    \label{fig:spreading}
\end{figure}

To project the particles onto the measurement manifold, we marginalize out the spatial dimensions measured by the sensor. 
For a DoA sensor, this means ignoring the angular position of the particles as measured from the agent, considering only their distance from the agent. 
We then order the particles based on this range, and the edges of the Voronoi cells are then computed by bisecting the distance between each pair of particles, and the centers of the Voronoi cells are found by bisecting the edges. 
Finally, for each particle we compute the signed distance from the particle to the corresponding Voronoi cell center. 
This distance is multiplied by the unit vector pointing from the agent to the particle to get the additional movement vector for that particle as illustrated in Fig. \ref{fig:spreading}. 

\subsubsection{Scaling \& Convergence}
Spreading particles along the measurement manifold helps align the particles with the uncertainty of the underlying distribution. 
However, if the particles are already converged on the target, we want to limit spreading. 
We can balance exploration and exploitation here by applying a scaling factor to the movement vectors before applying them to the particles. 

During periods of inter-agent dropout, we want to maximize spreading, so the scaling factor is set by default to a value of 1. 
However, while inter-agent communications are frequent, we can use some measurement of convergence to scale the spreading. 
We define the \emph{inter-set dispersion matrix} $E$ as the second-moment matrix of pairwise displacements between particle sets $\mathcal{P}^i_t$ and $\mathcal{P}^j_{t,s}$:

\begin{equation}\label{eq:E}
E = \frac{1}{MN} \sum_{k \in M} \sum_{l \in N} (p^i_{t,k} - p^j_{t,l}) (p^i_{t,k} - p^j_{t,l})^T
\end{equation}

We use the largest eigenvalue of this matrix $\lambda_{max}(E)$ as our convergence metric for scaling the particle spreading.

\subsection{Geometric Singularity}
If the target passes directly between the two agents, the measured lines of bearing become parallel, and the two DoA measurements do not contain enough information to resolve the location of the target. 
As the target approaches this geometric singularity, the estimate becomes more sensitive to noise in either of the measurements. 

It is useful to quantify this singularity to compare estimation performance. 
Let $p_{true}$ denote the true location of the target, and $\theta^i_{true}, \theta^j_{true}$ denote the true bearing angles from each agent to the target.
We define the Jacobian 

$$ J(p) = \begin{bmatrix} \dfrac{\partial \theta_1}{\partial x} & \dfrac{\partial \theta_1}{\partial y} \\[4pt] \dfrac{\partial \theta_2}{\partial x} & \dfrac{\partial \theta_2}{\partial y} \end{bmatrix},$$

and take $\sigma_{\min} (J(p))$ to be its smallest singular value. 
We then define the singularity score as:

\begin{equation} \label{eq:sing_dist}
s(p) = \tanh\big(\sigma_{\min}(J(p))\big)
\end{equation}

The singularity score is small when the target is nearly co-linear with agents and approaches 1 as the agent-target geometry become well-conditioned for localizing the target.
We can use this score to determine the difficulty of a given estimate and break results into two categories, the nominal case where the agents should theoretically be able to track the target with little difficulty, and the singular case where where the agent is co-linear with the agents, and therefore we would expect the tracking estimate to deteriorate.

\subsection{Distributed Bayes Filter}
To provide a baseline for comparison, we also implement a distributed Bayes filter to estimate the target's position. 
The Bayes filter divided the estimation area into hexagonal tiles with a side length of one meter and tracked the probability of the target being in each tile.
The target's motion is modeled by diffusing the tile probabilities at each timestep according to some behavioral model.  
Measurement updates are done using the same method as the particle filter, calculating likelihoods using tile centers rather than particle locations. 
Consensus messages consist of the probabilities for each tile, which the receiver multiplies with its own probabilities. 
The resulting distribution is normalized to become the updated estimate distribution. 
At each timestep the Bayes filter calculates the average of the tile centers, weighted by the tile probabilities, and publishes this as the estimate of the target's position. 

Because the Bayes filter is discrete, its performance is bounded by the size of its tiles. 
On the other hand, the particle filter operates in continuous space. 
To allow for direct comparison between the filters, we converted the continuous particle estimate into a discrete tiled estimate by calculating the probability of each tile from the number of particles it contained. 
The particle filter's estimate was then computed by taking the weighted average of the tile centers, identically to the Bayes filter. 

\section{Experiments}
We tested our distributed particle filter algorithm in a field test at a lake using SeaRobotics HYCAT USVs for both sets of vehicles.
For this experiment, we utilized two agents and a target (Fig. \ref{fig:testing}).
The vehicles communicated with each other and a base station by sending ROS messages over a WiFi network set up at the lake. 
A central command node running on the base station controlled the experiment by sending mission parameters and vehicle commands. 
Each vehicle was equipped with station-keeping and waypoint controllers to execute position commands. 

Each vehicle obtained a state estimate from an onboard VectorNav VN-300 INS and published it as ground truth data over the network. 
The two agents obtained a bearing to the target using a simulated DoA sensor derived from the ground truth state estimates with added noise. 
Each agent ran a copy of the distributed particle filter algorithm, which took in DoA measurements and output consensus messages and estimates of the target's location.  
Inter-agent communication dropout was simulated by ignoring consensus messages from other agents when a dropout command was active.

\subsection{Missions}
\begin{figure}[t]
    \centering
    \includegraphics[width=\linewidth]{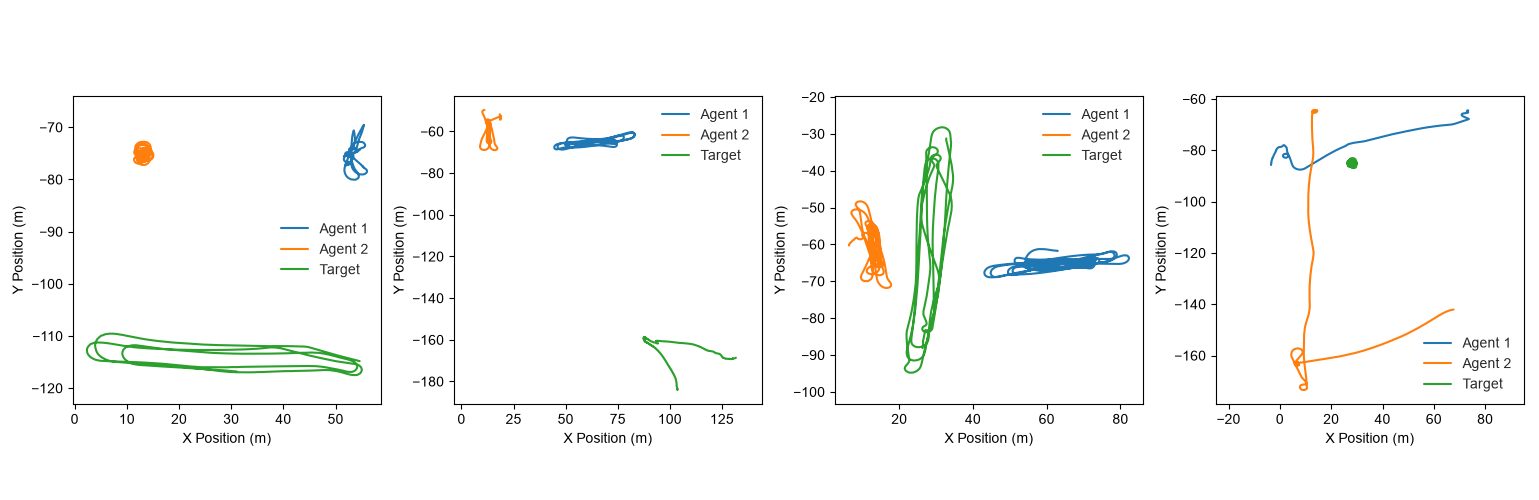}
    \caption{Agent and target trajectories from the four field test missions.}
    \label{fig:profiles}
\end{figure}

We tested the algorithm's performance across four missions that represent different scenarios. 
In Mission 1, the agents are still and the target moves back and forth in front of the agents. 
In Mission 2, the target moves in an area much further form the agents, where the singularity score is consistently low.
In Mission 3, the target travels directly between the agents to intentionally create geometric singularities. 
The agents move back and forth slightly to create variety in the geometry. 
In Mission 4, the target is stationary while the agents move around it.
The ground truth tracks based on GPS and inertial sensing are given in in figure \ref{fig:profiles}.

\subsection{Data Processing}
During the experiment, data was collected and each agent estimated the target location using the methods discussed, however the dropout was not included in order to generate the full data.
We then replayed the data with estimation algorithms and artificially induced dropout to test the effectiveness of the particle spreading proposed in this paper.
This allowed for the data to be reused, and for comparisons to be made between the methods using the same baseline data.

\section{Results}
We evaluated the Bayes filter, the particle filter without the particle spreading heuristic, and the particle filter with the spreading heuristic.
The three filters were first evaluated on their general performance across all four missions without artificial dropout. 
We then divided the missions into singular regions and nominal regions based on the singularity score, with sections that scored below 0.33 considered singular, and compared the performance in each category. 
Finally, we introduced varying amounts of artificial inter-agent communication dropout and observed the effect on performance. 

\subsection{Performance Across All Missions}

Table \ref{tab:error} reports the root mean squared error (RMSE) of each filter across all four missions. 
This includes missions 2 and 3, where singularity score was often low.  
The performance from all three filters is nearly identical, with a difference of only 0.11 meters between the best and worst filter. 
This difference is well below the standard deviation, making it statistically negligible. 
The standard deviations were also similar across filters and were all near the tile radius of 1m, which acts as a lower bound on filter performance. 
It is likely that using a smaller tile size would result in an even better approximation for all three filters. 
The particle filters likely would have performed better if their estimates were decoupled from the tiling.  

\begin{figure}[t]
    \centering
    \includegraphics[width=\linewidth]{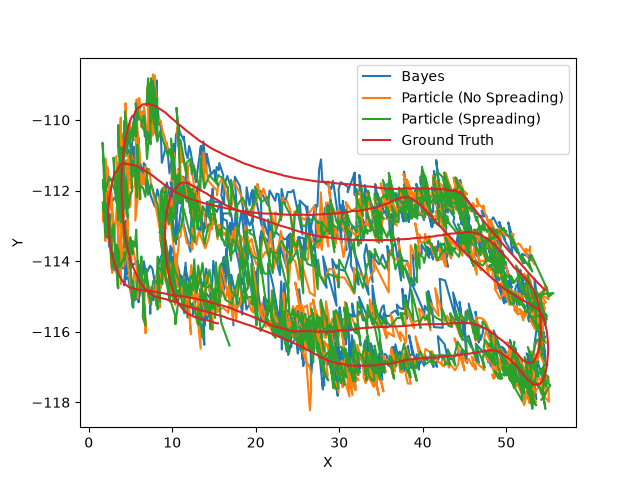}
    \caption{Target position estimates from Agent 1 on Mission 1.}
    \label{fig:estimate}
\end{figure}

This similarity between filters is illustrated in Fig. \ref{fig:estimate}, which shows the tracking results from Agent 1 on Mission 1. 
The filters deviate from the true target location at similar points and maintain a similar noise level. 

\begin{table}[htbp]
    \centering
    \caption{Error metrics across four missions.}
    \label{tab:error}
    \begin{tabular}{cccc}
        \toprule
        \textbf{Estimation Algorithm} & \textbf{RMSE} & \textbf{Std Dev} & \textbf{Max Error} \\
        \midrule
        Bayes                   & 1.590 & 1.135 & 14.713 \\
        Particle (No Spreading) & 1.470 & 1.122 & 14.968 \\
        Particle (Spreading)    & 1.519 & 1.175 & 15.204 \\ 
        \bottomrule
    \end{tabular}    
\end{table}

\subsection{Effect of Geometric Singularity on Performance}

The four missions were very different in terms of geometric singularity. 
Missions 1 and 4 had generally high singularity scores over 0.5, while mission 2 had a consistently low singularity score and mission 3 had frequent instances of full singularity. 
To better understand each filter's performance, we assign a singularity threshold of 0.33, classifying sections of each mission with singularity scores below this threshold as singular and the rest as nominal. 

\begin{figure}[t]
    \centering
    \includegraphics[width=\linewidth]{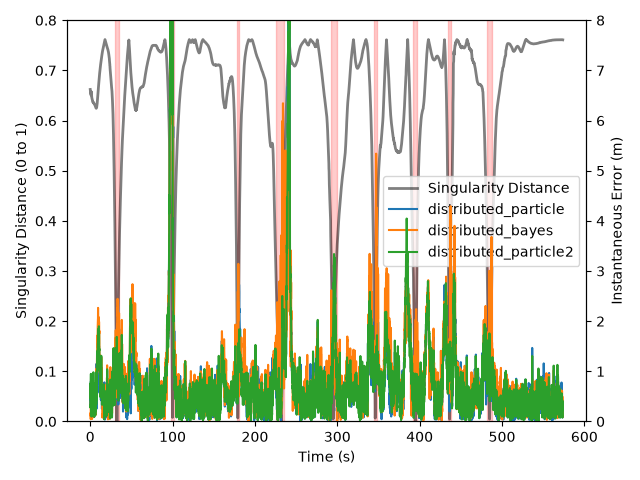}
    \caption{Agent 1 estimation error compared with singularity score for Mission 3.  The label distributed\_particle2 refers to the test with spreading.}
    \label{fig:singularity}
\end{figure}

Table \ref{tab:singularity} shows the RMSE for the singular and nominal zones. 
Once again, the performance between filters was very similar, with each filter performing more than twice as well in nominal zones compared to singular ones. 
We can see this in Fig. \ref{fig:singularity}, which shows the error over time for Agent 1 in mission 3. 
The error clearly spikes when the singularity score decreases, corresponding to the target passing directly between the agents. 
However, each agent successfully recovered quickly from each singularity. 

Table \ref{tab:singularity} also reports the Spearman correlation coefficient, which describes how well two variables (error and singularity score) can be described by a single monotonic function. 
A perfect correlation between the two variables would give a Spearman correlation of -1, meaning a decrease in singularity score perfectly correlates to an increase in error. 
The Spearman coefficient hovers near -0.5 for each filter, indicating a clear connection between singularity score and error.  

\begin{table}[htbp]
    \centering
    \caption{Error metrics across four missions.}
    \label{tab:singularity}
    \begin{tabular}{cccc}
        \toprule
        \textbf{Algorithm} & \textbf{Spearman} & \textbf{Nom. RMSE} & \textbf{Sng. RMSE} \\
        \midrule
        Bayes                   & -0.507 & 1.080 & 2.684 \\
        Particle (No Spreading) & -0.454 & 1.022 & 2.448 \\
        Particle (Spreading)    & -0.486 & 1.019 & 2.582 \\
        \bottomrule
    \end{tabular}    
\end{table}

\subsection{Effect of Dropout on Performance}

\begin{figure}[t]
    \centering
    \includegraphics[width=\linewidth]{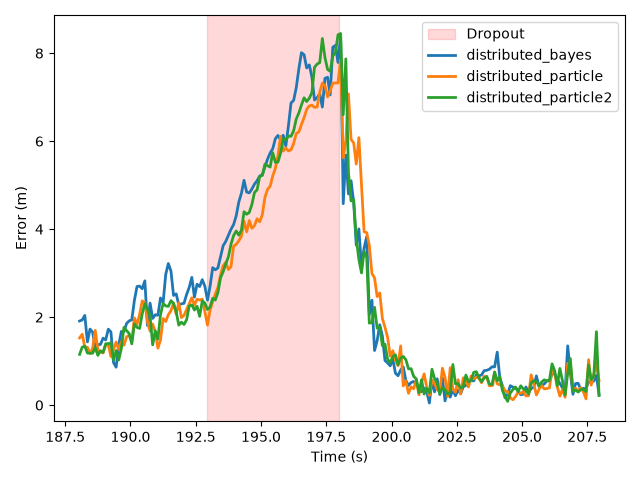}
    \caption{An example of estimation error rising during a communication dropout. The label distributed\_particle2 refers to the test with spreading}
    \label{fig:dropout_example}
\end{figure}

To test target tracking performance in periods of inter-agent communication dropout, an artificial dropout singal was injected for a set duration multiple times throughout mission 1. 
While the signal was active, agents ignored any received consensus messages. 
During this time, each agent only had its own DoA measurement, which is not sufficient to track the target, so estimate error accumulated. 
After the dropout signal was removed, agents resumed communication and the error returned to normal levels. 
This process is illustrated for one dropout event in Fig. \ref{fig:dropout_example}. 

\begin{figure}[t]
    \centering
    \includegraphics[width=\linewidth]{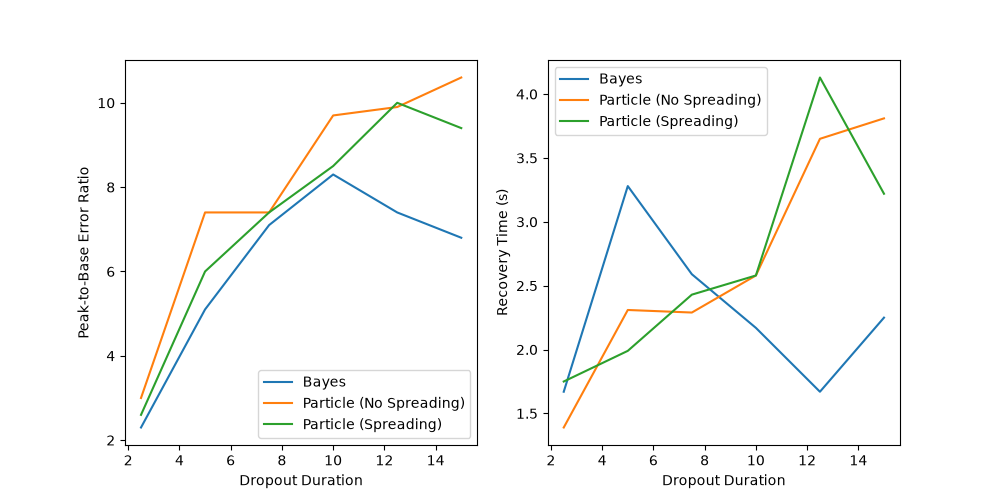}
    \caption{Mean peak-to-base error ratio and mean recovery time for different dropout duration.}
    \label{fig:dropout_stats}
\end{figure}

For each period of dropout, the mean error in the five seconds preceding signal activation became the baseline error. 
We recorded the peak error during the dropout period and the time it took for the estimate to recover, defined as the error dropping to within 1.1x the baseline error. 
We recorded these metrics for dropout duration from 2.5 seconds to 15 seconds, in intervals of 2.5 seconds. 
The results are shown in Fig. \ref{fig:dropout_stats}. 

As expected, the mean peak-to-base error ratio for each filter rose with increasing dropout duration. 
The rate of ratio growth seems to decrease at higher dropout duration, indicating that at some point, the particles (or tile probabilities) have spread enough that they are only loosely correlated with the target's position and will not get much worse.
The recovery time similarly increases at higher duration, though the effect is less predictable.

The Bayes filter outperforms both particle filters in both error ratio and recovery time.
This is not surprising; if we consider each tile center as a particle, then there is always some amount of particle coverage over the whole space, whereas the particle filters will struggle to track areas where they do not currently have particles. 
This gives the Bayes filter an inherent advantage when dealing with dropout, though this is balanced by its minimum error being bounded by the size of its tiles.

The particle spreading heuristic consistently reduced peak error in the particle filter by a small amount. 
Additionally, it better tracked the Bayes' filter performance than the unaugmented particle filter.
The filter with spreading also never performed significantly worse than the standard filter, and always outperformed it marginally in recovery time.
This indicates that the heuristic is not detrimental to the filter performance in the nominal case, and does lead to improved performance in certain edge cases.


\section{Conclusion}
We propose a novel augmentation for distributed particle filters in distributed sensor fusion problems and applies a spread along the sensor manifold when communications between agents are dropped.
This spreading de-emphasizes prior measurements from external agents that are outdated, focusing the estimate on the locally available data until new data from other agents is received.
This augmentation is demonstrated on experimental data gathered form real-world tests on USV platforms, and is shown that the performance is comparable to that of an no-spreading filter in the nominal case, and that the spreading filter outperforms the no-spreading one in certain edge cases where communications are dropped out.
These particle filters are also shown to have comparable performance to a baseline Bayes filter.

To continue this research we will explore the implementation of this spreading mechanism in the particle filter.
Particularly, we will further explore the kinds of edge cases in which this spreading mechanism may improve filter performance by examining delays of different duration, examining potential target behaviors during the dropout period, and tuning the various weights and parameters.
We also will explore how these effects may be extended to larger teams of agents featuring heterogeneous sensing capabilities.

\section*{Acknowledgments}
We thank the Naval Research Engineering Internship Program for providing the opportunity for C. N. to engage in this research. 

\bibliographystyle{IEEEtran}
\bibliography{references/library}

\end{document}